\documentclass{sig-alternate}

\usepackage{algorithm}
\usepackage{algorithmic}
\usepackage[utf8]{inputenc}

\begin{document}

%\conferenceinfo{GECCO'15,} {July 11-15, 2015, Madrid, Spain.}
%    \CopyrightYear{2015}
%    \crdata{TBA}
%    \clubpenalty=10000
%    \widowpenalty = 10000

\title{Optimal Neuron Selection: NK Echo State Networks for Reinforcement Learning}

\toappear{}

%\author{
%\begin{tabular}{ccc}
%Darrell Whitley & Renato Tinós & Francisco Chicano \\
%Department of Computer Science & Department of Computing and Mathematics & E.T.S. Ingeniería Informática \\
%Colorado State University & University of S\~ao Paulo & University of Malaga, Andalucía Tech, Spain \\
%Fort Collins, CO, USA & Ribeir\~ao Preto, SP, Brazil & University of Malaga, Andalucía Tech, Spain \\
%{whitley@cs.colostate.edu} & {rtinos@ffclrp.usp.br} & {chicano@lcc.uma.es} \\
%\end{tabular}}
%Darrell Whitley$^1$, Renato Tinós$^2$, and Francisco Chicano$^3$ \\
%\\
%$^1$University of M\'alaga, M\'alaga, Spain\\
%$^2$Colorado State University, CO, USA \\
%$^2$Colorado State University, CO, USA
%}
\numberofauthors{3} 
\author{
% 1nd. author
\alignauthor
Darrell Whitley\\
       \affaddr{Department of Computer Science}\\
       \affaddr{Colorado State University}\\
       \affaddr{Fort Collins, CO, USA}\\
       \email{whitley@cs.colostate.edu}
\alignauthor
Renato Tin\'os\\
       \affaddr{Department of Computing and Mathematics}\\
       \affaddr{University of S\~ao Paulo}\\
       \affaddr{Ribeir\~ao Preto, SP, Brazil}\\
       \email{rtinos@ffclrp.usp.br}
% 3nd. author
\alignauthor
Francisco Chicano \\
       \affaddr{E.T.S. Ingeniería Informática}\\
       \affaddr{University of Málaga, Andalucía Tech, Spain}\\
       \email{chicano@lcc.uma.es}
}

\maketitle
\begin{abstract}
This paper introduces the {\em NK Echo State Network}.
The problem of learning in the NK Echo State Network is reduced to 
the problem of optimizing a special form of a Spin Glass Problem known
as an NK Landscape.  No weight adjustment is
used;  all learning is accomplished by spinning up (turning on) or spinning
down (turning off) neurons in order to find a combination of neurons 
that work together to achieve the desired computation. 
%New highly efficient search methods now exist for optimizing NK Landscapes.
For special types of NK Landscapes, an exact global solution can be
obtained in polynomial time using dynamic programming. 
The NK Echo State Network is applied to a reinforcement learning 
problem requiring a recurrent network: balancing two poles on a cart 
given no velocity information.  Empirical results shows that 
the NK Echo State Network learns very rapidly 
and yields very good generalization.
\end{abstract}

%\category{1.2.8}{Artificial Intelligence}[Problem Solving, Control Methods, and Search]

%\terms{Algorithms, Theory}

\keywords{Neuroevolution, Echo State Networks, Neural Networks, NK Landscapes, Reinforcement Learning}

\section{Introduction}
\label{sec:int}

This paper introduces the {\em NK Echo State Network}.
Enhancements are added to the Echo State Network such that
the problem of learning is reduced to the problem of
optimizing an NK Landscape.  
The binary input to the NK Landscape turns on and off a set 
of $N$ neurons in the enhanced Echo State Network.

The dominant learning paradigm for training neural networks 
involves adjusting the weights that connect
artificial neurons using back propagation \cite{Rummelhart1995}. 
Similar learning methods
are also used by support vector machines:  for
classification problems, both methods involve optimizing 
vectors of weights which create and move interacting hyperplanes 
to produce decision surfaces that can be used to
separate training examples into the appropriate classifications.  
%Support vector machines provide additional guarantees about the
%margins around the hyperplanes and the separation between
%positive and negative classification examples.

``Reservoir computing" refers to a class of neural
network learning models, of which 
Echo State Networks is one example. 
Reservoir computing networks use a reservoir of sparsely connected
artificial neurons that have randomly generated weighted connections.
The input neurons are connected to neurons in the reservoir, and the output neurons
also are connected to neurons in the reservoir.  The weights inside of the
reservoir of neurons are not adjusted by learning.  Thus, 
reservoir computing has moved one step away from optimizing
vectors of weights as a learning paradigm; its ability to learn
depends on having a large reservoir of potentially useful neurons, and also,
having a way to determine which neurons are useful.   
In reservoir computing, learning still involves adjusting the 
weights that connect artificial neurons in the reservoir to the outputs
\cite{lukovsevivcius09}
\cite{lukovsevivcius12}
\cite{principe07}.

In the current paper, the problem of learning is 
recast as a ``Spin Glass Problem" in which
N neurons interact with K other neurons, and
each neuron spins up (on) or down (off) 
in order
to find a combination of neurons that work together
to achieve the desired computation.  The limitation
to K interactions can make the 
optimization problem tractable.

For the NK Echo State Network, there exists new highly
efficient optimization methods for that make it possible 
to know in constant time exactly which neuron to turn on or off in order
to obtain an improvement in performance \cite{ChicanoW2014}.
For special classes of NK Echo State Networks, we can prove 
convergence to a combination of neurons that is guaranteed 
to be globally optimal; an exact solution to the problem of
determining which neurons to use can be obtained 
in polynomial time using dynamic programming \cite{wright2000}.

The class of $k$-bounded pseudo-Boolean functions
are defined such that 1) the problem representation
(i.e., the domain) is the set of binary strings, 
and 2) the output of the evaluation function (i.e., the co-domain) is the set
of real numbers.  Pseudo-Boolean problems for inputs of length $N$ are $k$-bounded when the evaluation function
can be decomposed into a linear combination of subfunctions, and each
subfunction takes a subset of at most $k$ bits as input, where $k$ is a constant. 
NK Landscapes \cite{kauffman93} \cite{TomassiniOchoa2008} 
Spin Glass Problems \cite{Young1984}
and MAX-kSAT \cite{Kirkpatrick1994critical}  \cite{Mitchell1992hard}
are examples of  
well known $k$-bounded pseudo-Boolean optimization problems. 
All of these problems use an evaluation function of the form:

\begin{equation} 
f(\mathbf{x})= \frac{1}{M} \sum_{i=1}^{M} f_i(\mathbf{x})
\end{equation} 
where $X$ is the function domain and $x \in X$ is a bit vector, 
and each subfunction $f_i$ is evaluated using a subset of $k$ bits drawn from the bit vector $x$.  
When $k$ is small, each function $f_i$ can be 
expressed as a lookup table. 
In MAX-kSAT, each subfunction is a clause in CNF form, where the size of the clause
is less than or equal to $k$.    Each clause is true or false 
(returning 0 or 1) and the evaluation function $f$ sums over all $M$ clauses. 
In an NK-Landscape, $M = N$  and each subfunction $f_i$ uses bit $x_i$ as well as 
$K$ additional bits as input.  Thus, for NK-Landscapes, $k=K+1$
and the output of subfunction $f_i$ can be any real valued number.

To construct an NK Echo State Network, we convert the problem of selecting
which neurons to utilize into an NK-Landscape optimization problem.
In the current paper, we will consider an enhanced Echo State
Network with a single output.  One of the enhancements is
to use an ensemble of $N$ outputs.  A second enhancement is
to add an additional layer of $N$ neurons between the reservoir
and the ensemble of outputs; we will refer to this layer of neurons
as the {\bf probe filter}.   Each neuron in the
{\bf probe filter} acts as a probe, sampling the reservoir 
so as to create a different neural circuit.  
Each of the $N$ neurons in the ensemble of outputs
is connected to $k$ neurons in the {\bf probe filter}.    
This use of the {\bf probe filter} has some similarities to the
use of filter neurons by Holzmann and Hauser in Echo State networks \cite{holzmann2010};
they use the filter neurons to add an additional layer of weight optimization.
However, we have the very different purpose for the filter neurons.

The ensemble of $N$ output neurons combined with the $N$ neurons 
in the {\bf probe filter} can be expressed as an NK Landscape optimization problem.   
Each of the neurons in the {\bf probe filter} layer is a neural circuit;  the 
NK-Landscape is agnostic about how the neural circuits are created and needs
no information about the reservoir. 
Each of the $N$ output neurons becomes a subfunction in the
NK-Landscape, and is connected to $k$ neurons (neural circuits) in 
the {\bf probe filter} layer;  a binary string of length $N$ turns on 
and off the $N$ neurons in the {\bf probe filter}.   

At most $2^kN$ online
learning samples are needed to convert a learning problem into 
an NK-landscape.  The NK Landscape function is a lookup table which
stores all of the $2^k$ evaluations for each of the $N$ outputs.  
The NK-landscape can then be optimized offline using highly efficient optimization methods.  
This optimization process selects the best subset of neural circuits
from the {\bf probe filter} layer to use for the assigned learning problem.  

The transformation of the neural network training problem into 
an NK Landscape optimization problem represents a novel approach
to learning.  Given current interest in ``Deep Learning" and
more complex forms of neural networks, this approach to learning
may have a wide range of applications. 

We illustrate this new learning method by applying it to the
reinforcement learning problem of balancing two poles on a cart while
providing only cart position, and the two pole angles as input.
This means the NK Echo State Network must learn to compute velocity
information.   Our empirical results shows that the NK Echo State Network
not only learns rapidly, the resulting networks also produce 
very good generalization.

\section{Background: Neural Networks and Reservoirs}
\label{sec:met}

In his 1987 book {\bf Neural Darwinism}, G.M. Edelman advanced the idea
that ``group selection" acting on neurons could result in
a computational form of learning.   Stated concisely,  a diverse set
of inputs can be used to test and select for neural circuits that
respond appropriately to those inputs.  For example, Edelman 
conjectured cell death (as well as cell duplication and the development
of cell axons and dendrites) might also be controlled by some form
of functional selection.   Reduced to its simplest form, this
gives rise to a theory where learning might be achieved in a neural
architecture by turning on and turning off neurons as part of an
effort to identify ``useful neural circuits."  

In the evolutionary computation community, there is a long history
of neuroevolution spanning 25 years.   This work combines evolutionary
optimization methods with neural networks and related machine learning
methods such as support vector machines and echo state machines.
Most of these methods utilize some mixture of learning the architecture
of the neural network (including the number of neurons to use and how
they should be connected) as well as how to learn the weights in the networks.

The dominant paradigm for training most neural networks is
back-propagation, where learning almost exclusively focuses on adjusting 
the weights in the neural network.    Much of the work in the evolutionary
computation community has focused on reinforcement learning applications.
These are often control problems as well.
In reinforcement learning applications the evaluation of the system is
based on defining the {\em desired behavior} of the system, because it 
is not possible to directly know the correct actions 
to take in order to control the system.
An additional problem
is that reinforcement learning problems are often time dependent and the
input data to the neural network are expressed as a time series.  This means
that ``recurrent neural networks" are often required for reinforcement learning
applications.  Thus, reinforcement learning applications pose two problems
for traditional gradient methods such as back propagation:
1) only the desired behavior of the system is specified, not
the specific desired actions that the systems should display in response to
a particular input, and 2) recurrent neural networks are not easily trained
using gradient methods such as back propagation. 

% Figure 1 %
\begin{figure*}[ht]
\centering
\epsfig{file=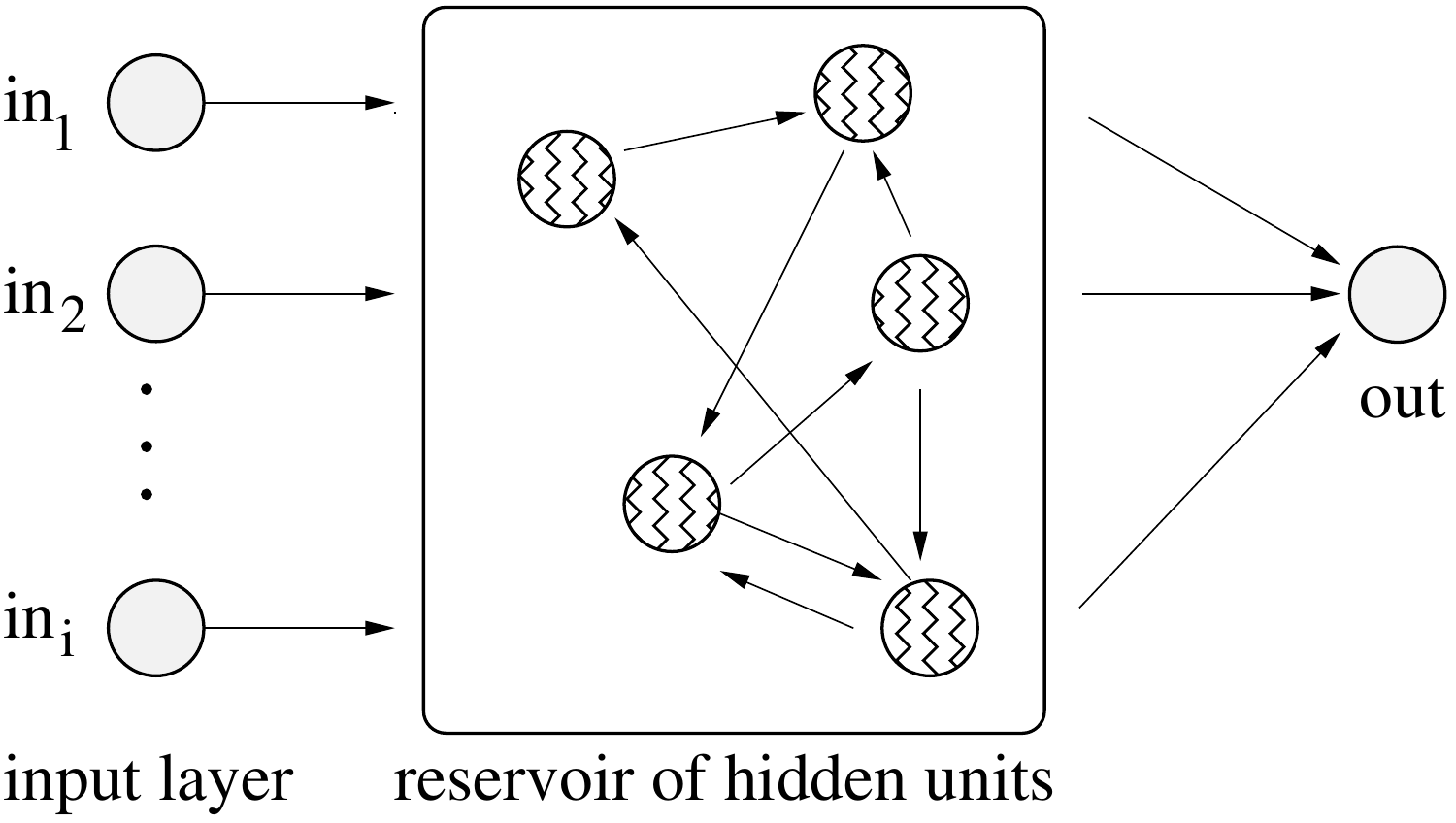, width=0.45\textwidth}
\caption{An Echo State Network with an input layer, a reservoir of neurons,
and a single output.  The recurrent neurons of the reservoir are sparsely connected with density $\alpha$.  There is a single output neuron in this case.  All neurons in the reservoir 
are connected to the output neuron. There are no recurrent connections originating from the output neurons.}
\label{fig:echo1}
\end{figure*}

In the last 10 years another approach to training recurrent neural networks for
reinforcement learning has focused on ``reservoir computing" methods, one example of
which are ``echo state networks."    A sparsely connected set of neurons 
with recurrent connections is created; the weights between this set of neurons
are set randomly, typically with random values that are relatively small 
(e.g. less than $\pm 1.0$).   This forms the reservoir.   Inputs neurons are
then connected to the neurons in the reservoir.   Neurons in the reservoir are
also connected to the output neurons.    This idea is illustrated in Figure 1,
where there is a single output.    One of the motivations for reservoir computing
is the work of Schiller and Steil \cite{schillerstein2005} which shows that when
gradient methods are used to training recurrent neural networks, most of the
weight changes occur in the weights that connect to outputs, even if the methods
are being used to change all of the weights in the network.    Thus, given a large
reservoir of neurons, some combinations of neurons should be more useful 
than others.  
Training methods
that optimize the weights between the neurons in the reservoir and the outputs
can, essentially, determine which neurons and groups of neurons are doing
useful computations.  In this sense,  reservoir computing has connections
back to Neural Darwinism and, more loosely, to neuroevolution. 

This reservoir computing model addresses the problem of dealing with
time dependent inputs that require recurrent neural networks.
But it does not directly address the reinforcement learning problem:
we still only know the desired behavior of the system, but not exactly 
what actions the system should display from one time step to the next.
Thus, traditional error propagation methods such as Back Propagation
cannot directly be  used to train Echo State Networks for
reinforcement applications.

One could use temporal difference methods or Q-learning methods for
reinforcement learning problems. However, \linebreak
Gomez, Schmidhumber and Miikkulainen \cite{gomez2008}
have shown that a wide range of methods based on these ideas do not
scale up and do not work well on more difficult reinforcement learning
problems.  They used Q-learning with a Multi-layer Perceptron that mapped
state-action pairs to Q-values.  They also compared to methods such
as Sarsa($\lambda$) with Case Based Function Approximators  
and Sarsa($\lambda$) with a Cerebellar Model Articulation Controller \cite{santamaria98}.
They concluded these methods were less effective and less efficient
for complex reinforcement learning tasks compared to neuroevolution
based methods such as NEAT, ESP and CoSyNE \cite{gomez2008}.

Methods such as NEAT \cite{stanley2002} emphasize neuron selection and refinement;
thus, NEAT (and HYPERNEAT) exploit a kind of Neural Darwinism.
Methods such as ESP \cite{gomez1999} and CoSyNE \cite{gomez2008} focus on weight optimization for
a fixed architecture.

This current paper takes the work on reservoir computing one step 
further:  we reconfigure the learning problem so that no weight adjustment
at all is used during learning.  Instead, learning is accomplish using
only neuron selection.  

\section{Implementation Details} 

In the current paper,  we make two enhancements
to the Echo State Network that enable the network to identify 
``useful neural circuits." 

We will work with an Echo State Network with one output.
The Echo State Network with one output has
been enhanced by adding an ensemble of $N$ outputs, all of which compute the same
output.   
The second enhancement is to add a {\bf probe filter} layer with $N$
neurons.  The $N$ neurons in the {\bf probe filter} layer in effect probe the
reservoir to generate $N$ different neural circuits.   Each neuron of the output 
layer is connected to $k$ neurons in the {\bf probe filter} layer.   This is illustrated
in Figure 2.  

% Figure 2 %
\begin{figure*}[ht]
\centering
\epsfig{file=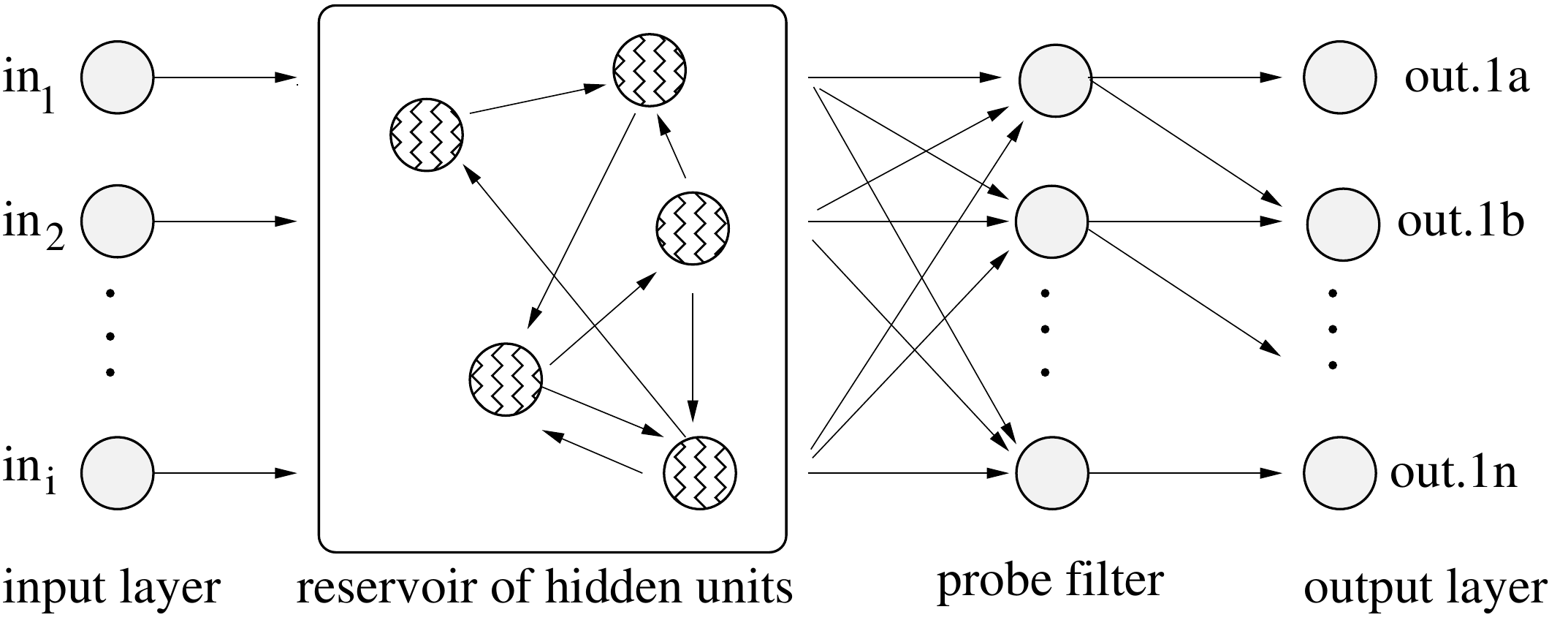, width=0.63\textwidth}
\caption{An NK Echo State Network.  The Echo State Network with one output has
been enhanced by adding an ensemble of $N$ outputs, all of which compute the same
output.  The second enhancement is to add the {\bf probe filter} layer.
Each neuron of the output layer is connected to $k$ neurons in the {\bf probe filter}
layer.  The neurons of the {\bf probe filter} layer are turned on or off according 
to a binary vector $\mathbf{x}$. If $x_i=1$, the $i^{th}$ neuron is turned on; 
if $x_i=0$, the $i^{th}$ neuron is turned off.  All weights are randomly created 
and are kept fixed during the optimization of $\mathbf{x}$.}
\label{fig:echoNK}
\end{figure*}

In the Echo State Network used here, all the weights of the network are randomly created 
and kept fixed during its training.  During the training, only the bit vector $\mathbf{x} \in \mathbb{B}^N$ is optimized;
the vector $\mathbf{x}$ specifies which neurons in the {\bf probe filter} layer are activated or not. 
If $x_i=1$, the $i$-th neuron of the {\bf probe filter} is turned on, i.e., its activation is used 
as input for neurons in the output layer connected to it.  If $x_i=0$, the $i$-th neuron of 
the {\bf probe filter} layer is turned off. 
The connectivity of the output layer to the {\bf probe filter} layer must be sparse. 
Each neuron of the output layer receives inputs from the outputs of $k$ neurons of the {\bf probe filter} layer. 
Following the usual definition of an NK-Landscape, the $i$-th neuron of the {\bf probe filter} layer 
is always connected to the $i$-th output neuron.   
In NK-Landscapes, each bit in each subfunction interacts
with K additional bits. 

We will use both the {\em adjacent} NK-Landscape model and the {\em random} NK-Landscape model
in our experiments.  In the adjacent NK-Landscape model, the $i$-th output neuron is connected 
to neurons $i$, $i+1$, ..., $i+K$ of the {\bf probe filter} layer.  
In the random NK-Landscape model, the $i$-th output neuron is connected to neuron $i$ and 
K other random neurons of the last hidden layer.   Solving the optimization
problem for the random NK-Landscape is NP-Hard.  However, a polynomial time
dynamic programming algorithm yields a global optimum for 
the adjacent NK-Landscape problem \cite{wright2000}.

%The Echo State Network is employed to learn to reproduce a function with 
%output $\psi \in \mathbb{R}$ and input vector $\mathbf{u} \in \mathbb{R}^{p} $. 

Our implementation will use two neuron selection mechanisms.  1) The function $mask(i,j)$ indicates 
an architectural feature of the neural network: it returns the $j^{th}$ neuron in the {\bf probe filter}
layer that provides an
input to the output neuron $i$.  This level of detail usually is not explicitly denoted in neural networks.  
But here it is an important part of the NK-Landscape design.   2) The vector $x$ 
turns on and off neurons in the {\bf probe filter} layer.   

At each iteration the $i^{th}$ output of the NK neural network is given by:
\begin{equation} 
y_i = \phi \Bigg( \sum_{j=1}^{K+1}  w_{mask(i,j),i}~ S(mask(i,j))~ x(mask(i,j))  \Bigg)
\label{eq:NKnnet_out}
\end{equation} 
where $\phi(.)$ is a sigmoidal shaped activation function. 
The function $mask(i,j)$ is a look-up table that returned the index of the $j^{th}$ neuron 
that connects to output $i$.  Assume $mask(i,j) = q$.  Thus, $w_{q, i}$ is the weight between the
$q^{th}$ neuron in the {\bf probe filter} which is also one of $K+1$ neurons that 
connects to the $i^{th}$ output neuron.
$S(q)$ is the output of neuron $q$ of the {\bf probe filter} layer; finally, $x(q)$ indicates
if the $q^{th}$ neuron in the {\bf probe filter} layer is currently turned on or off.  
If $x(q)=1$ then $w_{q,i}*S(q)*x(q) =  w_{q,i}*S(q)$. 

To see how this relates to standard neural networks, assume we defined an alternative neural network
where all neurons in the {\bf probe filter} are connected to {\em every} output neuron, but only $K+1$ of the
weights that connect to a specific output neuron are non-zero.  
Thus a zero weight is the same as no connection.  
Let the alternative weight vector be denoted by $w'$. 
Note that the original network and the alternative network yield identical computations. 
Again, assume that $mask(i,j)=q$.  
\begin{eqnarray*} 
\label{eq:NKnnet_out2}
y_i & = & \phi \Bigg( \sum_{j=1}^{K+1}  w_{mask(i,j),i} S(mask(i,j)) x(mask(i,j))  \Bigg) \\
       & = & \phi \Bigg( \sum_{q=1}^{N}  w'_{q,i} S(q) x(q) \Bigg) 
\end{eqnarray*} 
Finally, assume that all of the bits in vector $x$ are set to 1 so that all of 
the neurons in the {\bf probe filter} layer are on: we then obtain a
standard neural activation function:

\begin{equation*} 
y_i  =  \phi \Bigg( \sum_{q=1}^{N}  w'_{q,i} S(q) \Bigg)
\end{equation*} 

\subsection{Mapping to an NK-Landscape}

In the neural network investigated here, every output neuron is evaluated independently. 
This is because we want the set of $N$ outputs to operate as an ensemble. 
Because the weights of the network are fixed, the evaluation of the $i^{th}$ output 
neuron for a given problem depends 
only on the binary vector $\mathbf{x}$ and on the function $mask(i,j)$.  

We denote the evaluation of the $i^{th}$ output neuron as $f_i(\mathbf{x})$;
we can assume function $f_i$ will automatically implement $mask(i,j)$ when
passed an input of length $N$.  It is also convenient to assume that
$f_i$ can also take an input of length $K+1$;  
thus $f_i(\mathbf{x}) = f_i(0101)$ if $f_i$ uses $mask(i,j)$ to extract the
bit pattern $0101$ from vector $x$.  

For example, assume $N=20$ and $K+1 = 4$.  Also assume that the ``on/off" pattern 
of the 4 neurons used by output neuron number 19 is given by 0101.
Then $f_{19}(\mathbf{x}) = f_{19}(0101)$ evaluates the behavior of the network at
output neuron 19 when then the 1st and 3rd neuron that feed into output
19 are turned off, and the 2nd and 4th neurons are turned on.  
A single number is stored representing the performance of output neuron 19
for using the first and third neuron.   For an output neuron $i$, the performance
is recorded for every possible ``on/off" pattern over the $K+1$ neurons
to which it connects in the {\bf probe filter}.  The results in a look-up
table of performance evaluations with $2^{K+1}$ entries for each subfunction $f_i(x)$.  

This is repeated for each of the $N$ output neurons.  Thus, the total number of 
Echo State Network evaluations will always be exactly $2^{K+1}N$ (unless there
is some overlap in the computation of the subfunctions that would
allow some of them to be evaluated in parallel).  

We optimize the function $\mathbf{x}$ in order to maximize $f_i(\mathbf{x})$ 
when summed across all of the $N$ outputs of the Echo State Network. 
Thus, for the combined $N$ outputs, the evaluation function is:

\begin{equation} 
\label{eq:ann_eval}
f(\mathbf{x})= \frac{1}{N} \sum_{i=1}^{N} f_i(\mathbf{x})
\end{equation} 

The evaluation function given by Eq.~\ref{eq:ann_eval} is exactly the same evaluation function used in the NK landscapes. 
In this way, optimizing the enhanced Echo State Network is equivalent to optimizing an NK landscape. 
%In NK landscapes, the individual fitness contributions to $f_i(\mathbf{x})$ are random. 
%In this case, the contributions are not random, but are the result of utilizing a random subset
%of neurons from the {\bf probe filter} layer. 

The algorithms for training the enhanced Echo State Network using NK landscape 
optimization methods are given in the following descriptions.

\vspace{0.1in}
\noindent 
\textbf{Algorithm 1}: 
\begin{description}
  \item[i.] Create an artificial neural network $ANN(\mathbf{M})$ with at least one hidden layer. 
The last hidden layer is the {\bf probe filter}.
The matrix $\mathbf{M}=[\mathbf{m}_1 \ldots \mathbf{m}_N] $ defines the connectivity between 
the neurons of the {\bf probe filter} and the neurons in the output layer. 

The connections between the ensemble of outputs and the {\bf probe filter} layer
are determined by $\mathbf{mask(i,j)}$ which has exactly $K+1$ ones (indexed by $j$) defined 
according to the NK neighborhood model (adjacent or random); the $K+1$ neurons selected by 
$\mathbf{mask(i,j)}$ feed into output $i$.   
All the weights and thresholds of the $ANN(\mathbf{M})$ are random. A vector $\mathbf{x} \in \mathbb{B}^N$ is used 
to turn on or off the neurons in the {\bf probe filter} layer.  
\item[ii.]  Evaluate each output 
neuron of $ANN(\mathbf{M})$ for all combinations of the elements of vector $\mathbf{x}$ used by the output neuron. 
Since each output is connected to $K+1$ neurons, there are $2^{K+1}$ combinations for each output.
Thus, we need $N 2^{K+1}$ total evaluations to evaluate all subfunctions $f_i(\mathbf{x})$. 
For example, if $N=6$, $K=2$ and the adjacent neighborhood model is used, then the
subfunction $f_i(x)$ uses 
$\mathbf{mask(i,j)}=[1,1,1,0,0,0]^\textrm{T}$. 
Thus, the first output is evaluated for 
all eight combinations of $x_1,x_2,x_3$:  000, 001, 010, ..., 110, 111. 
  \item[iii.] Save $ANN(\mathbf{M})$. Also, save $\mathbf{M}$ and 
the individual performance evaluations $f_i(\mathbf{x})$ for all combinations 
of the elements of vector $\mathbf{x}$ used by the output neuron.
\end{description}
\vspace{0.01in}

\vspace{0.1in}
\noindent 
\textbf{Algorithm 2}: 
\begin{description}
  \item[i.] Load $\mathbf{M}$ and all of the $N2^{K+1}$ subfunctions corresponding to $f_i(\mathbf{x})$.
  \item[ii.] Create an NK model with $\mathbf{M}$ and the subfunctions $f_i(\mathbf{x})$.
  \item[iii.] Optimize the NK model by optimizing the vector $\mathbf{x}$. 
   \item[iv.] Save the best decision vector $\mathbf{x}^*$. 
\end{description}
\vspace{0.01in}

\vspace{0.1in}
\noindent 
\textbf{Algorithm 3}: 
\begin{description}
  \item[i.] Load the artificial neural network $ANN(\mathbf{M})$ created 
in Algorithm 1 and the best decision vector $\mathbf{x}^*$ found in Algorithm 2.
  \item[ii.] Evaluate $ANN(\mathbf{M})$ with vector $\mathbf{x}^*$. 
The signal employed to reproduce the output of the mapped function 
is composed by a weighted combination of 
the outputs $y_i(\mathbf{u}(t),\mathbf{x}^*,t)$, $i=1,\ldots,N$, of the neural network.
\end{description}
\vspace{0.01in}

\section{The Experiments}
\label{sec:res}
In order to test the methodology proposed in this work, the NK Echo State Network 
is applied to the double pole balancing problem without velocity information \cite{wieland91}.   
Two poles of different length are attacked to a cart that moves
on a track of fixed length.  Both poles are balanced by pushing
the cart to the left or to the right.  The force of the push is
allowed to vary.  
In this problem, the inputs of the artificial neural network 
at step $t$ are composed by scaled cart position and angles of the 
two poles at step $t$, i.e.,  the input 
vector 
 \[ \mathbf{u}(t)=[x_c(t)/x_c^{max}, \theta_1(t)/\theta_1^{max}, \theta_2(t)/\theta_2^{max}]^\textrm{T}  \]
where $x_c(t)$ is the cart position, $\theta_i(t)$ is the angle of the $i$-th pole, 
and $x_c^{max}$ and  $\theta_i^{max}$ are the maximum allowed values used to scale the inputs between -1 and +1. 
All neurons use the hyperbolic tangent function with outputs between -1 and +1
as the sigmoidal squashing function.    Since velocity information is
not given as input,  the recurrent Echo State Network is needed to learn 
to estimate velocity.  

In Algorithm 1, each output of the network is independently evaluated. 
The action (in Newtons) applied to the cart 
at iteration $t$ when evaluating the $i$-th output for solution $\mathbf{x}$ is given by:
\begin{equation}
action(t)=10y_i(\mathbf{u}(t),\mathbf{x},t)
\end{equation}
The following objective function was introduced by Gruau, Whitley and Pyeatt \cite{gruauwhit96}
and has been regularly used for the problem of balancing two poles on
a cart by a significant number of researchers \cite{stanley2002,durr2006,jiang08}.
\begin{equation}
\label{eq:fit_func}
f=0.1f_1 +0.9f_{stable}
\end{equation}
where $t$ is the number of steps inside the success domain until a limit of 5000 steps and 
\begin{equation}
f_1= t/t_{max}  ~~~~~~~~~~~~\mbox{(Typically $t_{max} = 1000$)}
\end{equation}
and
\begin{equation}
f_{stable} \hspace{-0.09cm} = \hspace{-0.09cm} \left\{ \begin{array}{ll}
       0, & \textrm{if $t<100$}\medskip\\
        \frac{0.75}{\sum_{i=t-100}^{t} (|x_c(i)|+|\dot{x_c}(i)|+|\theta_1(i)|+|\dot{\theta}_1(i)| )}, & \textrm{otherwise},\\
        \end{array} \right.
\vspace{0.2cm}
\end{equation}
The track length is given by $x_c \in [-2.4, 2.4]$ meters; beyond this range the
cart crashes into the ends of the track.   The system must keep both poles 
within $\theta_i \in [-36, 36]$ degrees of vertical. 
The function $f_1$ indicates that the cart and pole system has avoided 
a failed state (where a pole falls, or the cart crashes) for $t$ time steps.  
However, for small values of $t_{max}$ ($t_{max}=1000$ has been standard) 
a bang-bang control strategy might be learned that causes the system to 
become more unstable over time; even if the controller avoids failure 
for $t_{max}$ time steps, the system will become more unstable and
eventually fail when the system is run for more than $t_{max}$ time steps.
The second function $f_{stable}$ indicates the stability of the system during the last 100 time steps if $t \geq 100$. 
A higher value of $f_{stable}$ means that the system is staying close to the idea state:   close to the
center of the track, with small pole angles close to vertical, and with low velocities which means the poles
and the cart are not rapidly changing from one extreme state to another.  The value $0.75$ was established
by tuning the original {\em Cellular Encoding} network, which was the first neural network to solve this
optimization task in 1996; the use of the 0.75 value is probably now meaningless,
but it persists for historical and comparative reasons.

For the evaluation of $f$ in Algorithm 1, the systems always starts from the state $x_c(0)=\theta_2(0)=\dot{x_c}(0)=\dot{\theta}_1(0)=\dot{\theta}_2(0)=0$ and $\theta_1(0)=4.5$ degrees. 
The parameters of the double pole system used here are \cite{durr2006}: mass of cart equal to 1 kg, mass of pole 1 equal to 0.1 kg, mass of pole 2 equal to 0.01 kg, length of pole 1 equal to 1 m, length of pole 1 equal to 0.1 m, coefficient of friction of the cart on the track equal to 0.0005, coefficient of friction of the poles equal to 0.000002. 
The 4th order Runge-Kutta method with integration step equal to 0.01 was used.

\begin{table*}[t!]
\scriptsize
\centering
\caption{
Results for experiments with the Adjacent and Random NK Landscapes are shown for N=20. 
And Evaluation greater than 0.10 means the pole was always balanced;  a higher evaluation
means the state of the system was closer to the ideal state. 
Generalization is tested over 625 initial states (e.g. 625 represents perfect generalization).
Best Single Output shows the performance over the 20 outputs in the ensemble.
Best-of-100 shows the best generalization achieved out of the 100 runs.
The  symbol ``$s$'' indicates a significant difference between the generalization of the Best Single Output and the Ensemble.}
\vspace{0.1in}
\begin{tabular}{c|c|c|lc|lc} \hline\hline
  &   &  Evaluation &  \multicolumn{4}{c}{Generalization Test} \\ 
       &     &   Best Single Output  & \multicolumn{2}{c|} {Best Single Output} & \multicolumn{2}{c} {Ensemble}\\ 
N=20	  & $K$ &	mean $\pm$ std					   & mean $\pm$ std & Best-of-100 & mean $\pm$ std & Best-of-100 \\
\hline
Adjacent   &2 &0.12 $\pm$0.01 &241 $\pm$128 &550 &229 $\pm$160 &525 \\ 
           &3 &0.12 $\pm$0.01 &231 $\pm$134 &604 &304 $\pm$154 (s) &525 \\ 
 	   &4 &0.12 $\pm$0.01 &244 $\pm$137 &525 &321 $\pm$151 (s) &525 \\ 
           &5 &0.13 $\pm$0.01 &235 $\pm$142 &525 &377 $\pm$126 (s) &575 \\ 
           &6 &0.13 $\pm$0.01 &227 $\pm$133 &544 &339 $\pm$144 (s) &575 \\ 
           &7 &0.13 $\pm$0.01 &253 $\pm$141 &550 &362 $\pm$137 (s) &588 \\ 
 	   &8 &0.14 $\pm$0.01 &236 $\pm$144 &550 &389 $\pm$110 (s) &575 \\ \hline
Random 	   &2 &0.12 $\pm$0.01 &243 $\pm$151 &623 &232 $\pm$152 &520 \\ 
           &3 &0.12 $\pm$0.01 &245 $\pm$131 &569 &298 $\pm$156 (s) &525 \\ 
 	   &4 &0.12 $\pm$0.01 &278 $\pm$142 &577 &323 $\pm$145 (s) &575 \\ 
 	   &5 &0.13 $\pm$0.01 &252 $\pm$145 &575 &318 $\pm$161 (s) &532 \\ 
           &6 &0.13 $\pm$0.01 &269 $\pm$142 &535 &348 $\pm$149 (s) &575 \\ 
 	   &7 &0.13 $\pm$0.01 &220 $\pm$145 &544 &369 $\pm$141 (s) &600 \\ 
 	   &8 &0.14 $\pm$0.02 &204 $\pm$148 &614 &377 $\pm$146 (s) &573 \\ 
   \hline\hline
\end{tabular}
\label{tab:ANN_res1}
\end{table*}

\subsection{Empirical Results}

We present the results for the NK Echo State Network presented in Figure 2 with $N=20$, i.e., 
the number of neurons in the output layer.
The number of neurons in the {\bf probe filter} layer was also $N=20$. 
We tested seven values for the connectivity degree between the last hidden layer and the output layer: $K={2,3,4,5,6,7,8}$. 
Both the random neighborhood NK Landscape model was tested, as well as
the adjacent neighborhood NK Landscape model. 
%We also looked at Echo State Network configurations with one and two reservoirs.
%When there are two reservoirs,  interconnections and recurrent connection 
%exist only between neurons in a single reservoir.  

In reservoir computing, there are recurrent connections between neurons. 
The connectivity density ($\alpha$) for each reservoir is 10$\%$, i.e., each neuron in 
the reservoir has recurrent connections to 10$\%$ of the neurons in the reservoir. 
All weights of the NK Echo State Network are fixed, being randomly generated between [-0.6 , 0.6]. 
After the initialization, 
the recurrent weights in each reservoir are scaled with a spectral radius equal to 0.95. 
All neurons use the hyperbolic tangent function as the activation function. 

The number of runs is 100. 
In each run, Algorithm 1 is used to compute the individual fitness contributions $f_i(\mathbf{x})$. 
Thus, the NK Echo State Network is generated and each output $y_i$ is evaluated using Eq.~\ref{eq:fit_func} for all possible combinations of the elements of vector $\mathbf{x}$ used by the $i$-th output neuron. 
Then, Algorithm 2 is used to obtain the best solution $\mathbf{x}^*$ for the evaluation given by Eq.~\ref{eq:ann_eval}. 

We can use various methods to optimize the NK landscape; for the
adjacent neighborhood landscape model, we can use we use dynamic programming
to find the global optimum in polynomial time.   
For the random neighborhood landscape model,  
we can use new fast Local Search algorithms \cite{WhitleyChen2012} \cite{ChicanoW2014} to 
search the radius $r$ neighborhood; this algorithm is able to identify
improve moves that are $r$ steps ahead in constant time.  
When $r>1$ many local minima are eliminated
that exist in the Hamming distance 1 local search neighborhood.  
When $N$ is small (e.g. $N=20$) we can also 
use exhaustive enumeration to find the global optimum. 

Finally, Algorithm 3 is employed to evaluate the NK Echo State Network with the combined 
outputs acting as an {\em ensemble}.  The output of the ensemble at step $t$ is given by:
\begin{equation} 
\label{eq:com_out}
y_{ensemble}(\mathbf{u}(t),\mathbf{x}^*,t)= \sum_{i=1}^{N} a_i(\mathbf{x}^*) y_i(\mathbf{u}(t),\mathbf{x}^*,t)
\end{equation} 
where $\mathbf{u}(t)$ is the input vector of the neural network 
at step $t$, $\mathbf{x}^*$ is the solution vector obtained by 
Algorithm 2, $y_i$ is the $i^{th}$ output of the neural network 
and the weight $a_i$ is given by 
\begin{equation} 
a_i(\mathbf{x}^*)= \frac{f_i(\mathbf{x}^*)}{\sum_{i=1}^{N} f_i(\mathbf{x}^*)}
\end{equation} 
i.e., the outputs with better results have higher weights. 
This amounts to a simple form of learning where the weights are
set once to combine the ensemble into a single output.
%We tested also the uniform case where $a_i(\mathbf{x})=\frac{1}{N}$ and 
%the case where only a percentage of the best outputs are used in Eq.~\ref{eq:com_out}, but the results were similar or worse. 
The action to the cart at step $t$ is: 
\begin{equation}
action(t)=10y_{ensemble}(\mathbf{u}(t),\mathbf{x}^*,t)
\end{equation}

The ensemble is evaluated using the generalization test proposed in \cite{whitley1994}. 
In this generalization test, the ensemble is evaluated 625 times, each time with different initial settings for cart position, cart velocity, pole 1 angle, and pole 1 velocity.  
The initial pole 2 angle and velocity are always zero. 
The combination of five different initial settings for each variable is considered: 5, 25, 50, 75, and 95\% of a reduced range of the variables. 
The reduced range is between -2.14 and 2.14 m for cart position, -1.35 and 1.35 m/s for cart velocity, -3.6 and 3.6 degrees for pole 1 angle, -8.6  and 8.6 degrees/s for pole 1 velocity.  
The evaluation of the generalization test is the number of times that the system remains in the success domain after 1000 steps.

Table \ref{tab:ANN_res1} shows the results of experiments where the adjacent and random NK models were considered. 
%Table 1 shows the results of experiments where the adjacent and random NK models were considered. 
The reservoir utilizes 60 neurons. 
%The results of an experiment with $r=2$ reservoirs, each one with 30 neurons, are still presented for the adjacent model.  
The results of the generalization test 
for $y_{ensemble}(t)$ are given in the column indicated as ``Ensemble'', while ``Best Single Output'' indicates 
the best single value of $f_i(\mathbf{x})$ among the $N 2^{K+1}$ possible 
performance values obtained in Algorithm 1.  
The results of the generalization test shows that on average the ensemble generally yields
better generalization than the best single output neuron. 
The column marked ``Best-of-100" shows the generalization of the best NK Echo State Network
out of the 100 that were generated.
% the best evaluation of Eq.~\ref{eq:fit_func} for the individual outputs of 
% the neural network using the best combination of the respective inputs for the output neuron, 
% are also shown in the Table (this result is denoted $max(f_i(\mathbf{x},\mathbf{m}_i)$).  
%Evaluations of $f$ (Eq.~\ref{eq:fit_func}) higher than 0.1 indicate that 
%the double pole system successfully balanced the poles for 1000 steps.

% Table 2   
\begin{table*}[t!]
\scriptsize
\centering
\caption{For these experiments the size of the ensemble is $N=100$, 
the adjacent model is used and dynamic programming
is used to guarantee convergence to the global optimum.  
When the results from the ensemble using only the best 20 outputs 
are statistically different from the two other distributions 
(as indicated by the Wilcoxon signed rank test), 
a symbol ``$s$'' is shown. Again, 625 represents perfect generalization.
Generalization increases as N and K increase.}
\vspace{0.1in}
\begin{tabular}{c|c|lc|lc|lc} \hline\hline
  &  Evaluation &  \multicolumn{6}{c}{Generalization Test} \\ 
  &   Best Single Output  & \multicolumn{2}{c|} {Best Single Output} & \multicolumn{2}{c|} {All 100} & \multicolumn{2}{c} {The ``Top 20"}\\ 
 $K$ &	mean $\pm$ std	& mean $\pm$ std & Best-of-100 & mean $\pm$ std & best & mean $\pm$ std & Best-of-100 \\
\hline
2 &0.13 $\pm$0.01 &239 $\pm$137 &525 &324 $\pm$115 &525 &451 $\pm$79 (s) &525 \\ 
3 &0.13 $\pm$0.01 &278 $\pm$141 &566 &396 $\pm$109 &525 &478 $\pm$56 (s) &586 \\ 
4 &0.13 $\pm$0.01 &238 $\pm$151 &574 &437 $\pm$84 &525 &490 $\pm$48 (s) &575 \\ 
5 &0.14 $\pm$0.01 &231 $\pm$147 &533 &433 $\pm$84 &525 &489 $\pm$54 (s) &599 \\ 
6 &0.14 $\pm$0.01 &260 $\pm$160 &625 &462 $\pm$69 &525 &494 $\pm$46 (s) &599 \\ 
7 &0.15 $\pm$0.01 &206 $\pm$150 &550 &468 $\pm$63 &575 &498 $\pm$48 (s) &612 \\ 
8 &0.16 $\pm$0.02 &183 $\pm$139 &545 &480 $\pm$57 &525 &504 $\pm$47 (s) &614 \\ 
   \hline\hline
\end{tabular}
\label{tab:ANN_100}
\end{table*}

From Table 1 we can see that Ensemble regularly produces generalization results that
are significantly better than the generalization results associated with the
``Best Single Output" neuron.   
The Wilcoxon signed rank test (at the 0.05 significance level) 
was used to compare the generalization results.  
We note that the
generalization results for the  ``Best Single Output" were much better than we expected
given that the only ``learning" that was done was to pick the best of $2^{K+1}$ 
configurations of neurons in the {\bf probe filter} layer.
Table \ref{tab:ANN_res1} also shows there is little or no difference between
using the random NK Landscape model and the adjacent NK Landscape model.
This means that the adjacent NK Landscape model provides sufficient diversity
in neural circuits while still being regular enough to be solved in polynomial time.

In terms of performance,  any Evaluation above 0.10 means that the poles are
being balanced for $t=1000$ steps.  In this sense, learning was successful
for all values of $K$.  The mean evaluation of the output neurons increased as
$K$ increased.  This means that the pole and cart have 
less change in position and velocity over time, and thus, the system is more stable.  
It is also clear that ``Generalization" improves as $K$ is increased.   
Recall that generalization tests are executed over 625 start states.  When $N=20$, 
the generalization results ranges from a mean of 229/625 when 
$K=2$ to 389/625 when $K=8$ for the Adjacent NK Landscape model.
A score of 625 would mean that the system was able to control the system from
every possible start state in the generalization test suite; results
in Table 1 show this is possible (K=6, Best Single Output, Best-of-100). 

The variance for the generalization is high, ranging
from a standard deviation of 110 to 160 for the Adjacent NK Landscape model.
The results are very similar for the Random NK Landscape model.
This high variance means that some NK Echo State Networks are very much above average and doing an 
extremely good job at generalization, while other NK Echo State Networks are
very much below average.  The column marked ``Best-of-100" shows that the best networks
can achieve very good generalization: a few of the ``Best-of-100" NK Echo State Networks
have generalization results above 600.

\subsection{Using a Larger Ensemble}

We next asked what happens when $N$ is increased to $N=100.$  
These results are shown in Table \ref{tab:ANN_100};  only the
Adjacent NK Landscape model is used.
All of the runs were again successful at balancing the two poles.
The difference lies in the generalization.   With $N=100, K > 3$
we now see the average generalization improve to over 400 out of the 625 
start states; in addition, the standard deviation reduces to below 85.

As has already been noted, generalization improves as either
$K$ or $N$ increases.   This of course has a cost, since constructing the
NK-Landscape has cost of $N2^{K+1}$.  
The results labeled ``All 100" in Table \ref{tab:ANN_100} 
includes all 100 outputs in the ensemble.

We also considered one more strategy to improve generation. 
This strategy included no additional evaluations.  After the
{\bf probe filter} has been optimized by turning on and off neurons,  
we checked the behavior of the 100 output nodes in the ensemble
on the evaluation function, and selected the ``Top 20" best
outputs to create the ensemble. 

These results are also shown in Table \ref{tab:ANN_100} and
are labeled the ``Top 20."  As the results show, this again 
increases generalization at virtually zero additional cost.

\begin{table}[h]
\scriptsize
\centering
\caption{Evaluation Results comparing the Adjacent Neighborhood NK-Landscape with different values of 
N and K.   The results are also compared to other results in the literature. All of the 2008 results
are from Gomez, et al. \cite{gomez2008};  these results sometimes include improvements over other earlier published
results for the same methods. The ``Top 20" outputs were used to create the ensemble when N=100.}
\vspace{0.1in}
\begin{tabular}{c|c|c} \hline\hline
           &                 &                    \\
           &  Number of      &                    \\
Algorithm  &  Feedward Steps &  Generalization    \\
\hline

        &       &   \\
CE 1996 \cite{gruauwhit96}     & 840,000      & 300   \\
ESP  1999 \cite{gomez1999}     & 169,000      & 289   \\
ESP  2008 \cite{gomez2008}     & 26,342    & Not Reported  \\
NEAT 2002 \cite{stanley2002}      & 33,184    & 286   \\ 
NEAT 2008 \cite{gomez2008}     & 6,929     & Not Reported  \\ 
%CMA-ES 2008    & 6,061     & N.A.  \\ 
CoSyNE 2008 \cite{gomez2008}   & 3,416     & Not Reported  \\  \hline \hline
      &     &     \\ 
N=20, K=3      & 320       & 304     \\ 
N=20, K=5      & 1,280     & 377     \\  \hline
      &     &   \\  
N=100, K=2, Top 20 & 800      & 450  \\ 
N=100, K=3, Top 20 & 1,600    & 487  \\
N=100, K=4, Top 20 & 3,200    & 490  \\
   \hline\hline
\end{tabular}
\label{tab:All}
\end{table}

\vspace{-12pt}

\subsection{Comparative Results}

Recent papers have focused more on the number
of evaluations needed to achieve ``successful learning" and less on
the generalization of the neural network.
However,  in the end it is generalization that is important if the learned
solution is going to be useful.  
In Table \ref{tab:All} various results are reported
from a number of papers published over the last 18 years for the
double pole balancing problem when no velocity information
is provided as input.    One can observe a dramatic improvement
in the ESP algorithm between the years 1999 and 2008.   Likewise NEAT
shows considerable improvement between 2002 and 2008.   

Methods such as ESP and CoSyNE depend on having
a predefined and fixed architecture; almost a decade of experience
solving the double pole balancing problem with no velocities surely makes
it easy to select a suitable compact neural network architecture such
that the problem is reduced to just learning the weights.  But this also
exploits more human experience in configuring the neural network.

Here,  we report results for the NK Echo State Network for
configurations that keep the number of evaluations under 4000.
Using just 320 evaluations, the NK Echo State Network with $N=20$ and $K=3$ yields
an average generalization of 304 successes from the 625 possible start
states.  This level of generalization is similar to results reported
in the earlier literature.    But the number of evaluations is
very low.     Better generalization was achieved by setting $N=100$ 
and $K=3$ and then selecting the ``Top 20" outputs to be included
in the ensemble.  This configuration used 1600 evaluations, but 
the NK Echo State Network was able to successful balance the double
pole from 478 of the 625 start states on average.   If generalization
is important,  the best networks can be selected
after multiple runs. 

Finally,  it should be noted that the NK Echo State Network does 
not use an architecture that is specific to this problem.  
Theoretically, exactly the same architecture 
could be used to make stock market predictions, or any other
learning application where an Echo State Network
might be used.  In the sense,  the same reservoir can be
reused to solve very different machine learning
problems by reconfiguring the {\bf probe filter} layer.  

\section{Conclusions}

The problem of training an enhanced Echo State Network has been
converted into the problem of optimizing a spin glass system in
the form of NK Landscape.
Learning is accomplished by turning on and off neurons in a 
{\bf probe filter} layer.  The {\bf probe filter} is a layer
of hidden units that is inserted between
the reservoir of the Echo State Network and the output neurons. 
The neurons in the {\bf probe filter} layer are not recurrent,
but are connected to recurrent neurons in the reservoir.
 
A well known reinforcement learning benchmark 
was investigated that requires a single output neuron was
used to test the NK Echo State network.
An additional enhancement of the Echo State Network was the 
use of an ensemble of N outputs, all of which attempt to learn
the same decision function.  The ensemble is combined to yield
a single output. 

Our empirical results show that the resulting ``NK Echo State Network"
is able to learn the control task of balancing two pole on a fixed
track with no velocity information.  Learning was faster than other
results that have been reported in the literature. 
Furthermore, generalization improved as N and K were increased.

The approach of using ``neuron selection" to learn instead of
adjusting weight vectors represents a novel approach to training
neural networks.    With the current interest in ``Deep Learning,"
this methods may find additional applications in training other
types of complex neural networks.

\section{Acknowledgments}
This work was supported by FAPESP and CNPq in Brazil, the Air Force Office of Scientific Research, Air Force Materiel Command, USAF (under grant FA9550-11-1-0088), project number 8.06/5.47.4142 of Universidad de Málaga in collaboration with the VSB-Technical University of Ostrava, UMA/FEDER FC14-TIC36, Universidad de Málaga, Campus de Excelencia Internacional Andalucía Tech and the Spanish Ministry of Education (grant CAS12/00274).
The U.S. Government is authorized to reproduce and distribute reprints for Governmental purposes.

%\bibliographystyle{abbrv}
%\bibliography{NKEcho}

\end{document}